\documentclass[10pt,twocolumn,letterpaper]{article}
\newcommand{\benchmark}{\textbf{CAPTCHA-X}}
\usepackage[pagenumbers]{cvpr} 
\usepackage{xcolor}
\usepackage{colortbl}
\usepackage{multirow}
\usepackage{array}
\usepackage{pifont}
\usepackage[utf8]{inputenc}
\usepackage{textgreek}
\definecolor{lightblue}{RGB}{230, 243, 247}
\definecolor{lightgray}{RGB}{255, 255, 240}
\definecolor{lightred}{RGB}{255, 230, 234}
\definecolor{lightgreen}{RGB}{220, 255, 220}
\definecolor{lightpurple}{RGB}{230, 230, 250}

\definecolor{cvprblue}{rgb}{0.21,0.49,0.74}
\usepackage[pagebackref,breaklinks,colorlinks,allcolors=cvprblue]{hyperref}
\usepackage{stfloats}
\usepackage{placeins}
\title{Reasoning under Vision: Understanding Visual-Spatial Cognition in Vision-Language Models for CAPTCHA}
\author{
Python Song\quad
Luke Tenyi Chang\quad
Yun-Yun Tsai \quad
Penghui Li \quad
Junfeng Yang\\
Columbia University\\
Department of Computer Science
}

\begin{document}
\maketitle
\begin{abstract}
CAPTCHA, originally designed to distinguish humans from robots, has evolved into a real-world benchmark for assessing the spatial reasoning capabilities of vision-language models.  In this work, we first show that step-by-step reasoning is crucial for vision-language models (VLMs) to solve CAPTCHAs, which represent high-difficulty spatial reasoning tasks, and that current commercial vision-language models still struggle with such reasoning.
In particular, we observe that most commercial VLMs (e.g., Gemini, Claude, GPT, etc.) fail to effectively solve CAPTCHA and thus achieve low accuracy($\sim$ \textbf{15.7\%}), but our findings indicate that requiring the model to perform step-by-step reasoning before generating the final coordinates can significantly enhance its solving accuracy, this underscoring the severity of the gap.  To systematically study this issue, we introduce \textbf{\benchmark}, the first real-world CAPTCHA benchmark with reasoning, covering seven categories of CAPTCHAs (e.g., Gobang, Hcaptcha, etc) with step-by-step action solutions, and grounding annotations. We further define five reasoning-oriented metrics that enable a comprehensive evaluation of models’ reasoning capabilities.  To systematically explore the boundary of reasoning, we propose a reasoning-centered agentic framework built on large vision-language models. The framework operationalizes the model’s reasoning into executable actions. It achieves state-of-the-art performance on five high-difficulty CAPTCHA types and attains an average accuracy of \textbf{83.9\%} across all seven categories in our benchmark, substantially surpassing existing baselines.
\end{abstract}    
\section{Introduction}
\label{sec:intro}


CAPTCHAs were originally introduced as a security mechanism to distinguish humans from machines~\citep{vonahn2003captcha}. Early text-based CAPTCHAs exploited the limits of OCR~\citep{wang2018captcha}, but advances in computer vision shifted them toward complex visual–spatial puzzles requiring spatial reasoning, 3D mental rotation, and multi-step inference~\citep{gao2021research,luo2025open}. This evolution transforms CAPTCHAs from perception tests into probes of higher-level cognition, serving both as defenses against automated attacks and as testbeds for machine reasoning~\citep{ding2025illusion}. Today, they stand as real-world benchmarks for evaluating spatial intelligence in vision–language models, combining perception, reasoning, and decision-making~\citep{liu2023visual}.

With the rapid progress of vision–language models (VLMs), existing CAPTCHA benchmarks suffer from several fundamental limitations. While Open CaptchaWorld~\citep{luo2025open} introduces reasoning-related difficulty metrics, it lacks reasoning annotations, preventing a comprehensive evaluation of models’ reasoning abilities. Meanwhile, many recent general solvers (e.g., Halligan) achieve strong performance by combining VLMs with auxiliary tools and finetuned model~\citep{teoh2025bot-hard}~\citep{deng2024oedipus} ~\citep{wu2025mca}, yet they do not explicitly incorporate reasoning. Besides, most other datasets only provide CAPTCHA images with corresponding answers (such as coordinates) and evaluate correctness by measuring whether the distance between predicted and ground truth values falls within an empirically set threshold. This mismatch often yields offline results that fail to reflect online performance and fail to capture the reasoning processes underlying successful CAPTCHA solving, as we will discuss in detail in \autoref{sec:3.1}. Ultimately, a central gap remains: no prior work has definitively answered whether reasoning itself is the key to solving CAPTCHA.

In this paper, we create the first real-world benchmark \benchmark\ with reasoning annotations and show evidence that reasoning is the key to solving CAPTCHAs.
Directly applying commercial VLMs to solve CAPTCHAs, especially highly difficult tasks, achieves only an accuracy of 15.7\%. underscoring severe deficits in spatial reasoning. 

Once reasoning is introduced, however, performance statistically significantly improves by an average of 38.75\% relative to the non-reasoning baseline.
To further explore this finding. Inspired by the SayCan ~\citep{saycan2022arxiv} in robotics, which proposed the key idea of combining high-level semantic reasoning from language models with the real-world executability of robotic behaviors to bridge reasoning and action, we explore whether this principle can be transferred to our CAPTCHA solving task. We further ask, what would happen if reasoning itself became the core of action? In our large-scale experiments, we observed that reasoning significantly improves model performance, but it still has limitations. The models sometimes produce flawed logic or misidentify grid structures and spatial positions. To systematically evaluate and extend the boundary of reasoning, we design a agentic pipeline using VLMs to solve these above issues. Based on this, we construct a reasoning-centered executable agent prototype that achieves state-of-the-art results across multiple CAPTCHA tasks and establishes a new direction toward reasoning-centered intelligence, where agents act because they reason.


Our contributions can be summarized as follows:

\begin{itemize}[noitemsep,nolistsep,leftmargin=10pt]
    \item We introduced \benchmark, the first real-world CAPTCHA benchmark with reasoning annotations. \benchmark\ covers seven challenges with authentic annotations, region-level acceptance zones, and  reasoning steps to systematic evaluation of reasoning capability for VLMs.
    \item Using \benchmark, we demonstrated the importance of reasoning for CAPTCHA solving and exposed severe deficits in existing VLMs' spatial reasoning capability.
    \item Experiments on our benchmark show that incorporating reasoning improves performance by 38.75\% relative to the non-reasoning baseline, and statistical analysis confirms the improvement is highly significant (p $<$ 0.001), providing the systematic evidence that reasoning fundamentally improves model accuracy. Furthermore, we discover a Reasoning Scaling Law, showing that model performance follows consistent scaling relationships with different reasoning metrics.
    \item To extend this finding into practice, we develop a reasoning-centered agent framework that grounds reasoning into executable intelligence, reaching 83.9\% average accuracy across seven CAPTCHA types and setting new state-of-the-art performance on five categories.
\end{itemize}

\begin{table*}[t]
\centering
\small  
\caption{CAPTCHA Benchmark Comparisons.}
\label{tab:captcha-compare}
\begin{tabular}{l>{\centering\arraybackslash}m{1.2cm}%
                >{\centering\arraybackslash}m{1.3cm}%
                >{\centering\arraybackslash}m{2cm}%
                >{\centering\arraybackslash}m{1cm}}
\hline
\multicolumn{1}{c}{\bf Benchmark}  &
{\bf Real world} &
{\bf Reasoning} &
{\bf Region Consistent} &
{\bf Scale}
\\
\hline
Open CaptchaWorld~\citep{luo2025open} & \ding{55} & \ding{55} & \ding{55} & 225 \\
Halligan~\citep{teoh2025bot-hard}     & \checkmark & \ding{55} & \checkmark & 2600 \\
OEDIPUS~\citep{deng2024oedipus}       & \checkmark & \ding{55} & \ding{55} & 300 \\
MCA-Bench~\citep{wu2025mca}           & \ding{55} & \ding{55} & \ding{55} & 180000 \\
\benchmark\ (Ours)                    & \checkmark & \checkmark & \checkmark & 1839 \\
\hline
\end{tabular}
\end{table*}
\section{Related Work}
\label{sec:related_work}

\textbf{CAPTCHA Evolution and Benchmarking.}  
Over two decades, CAPTCHAs evolved from distorted text~\citep{vonahn2003captcha} to image-based challenges like Asirra, later broken by machine learning~\citep{hitaj2020capture}. This fragility spurred variants requiring logical reasoning and multi-step interaction. Recent benchmarks such as MCA-Bench~\citep{wu2025mca} and Bot-Hard~\citep{teoh2025bot-hard} emphasize multimodal reasoning and robustness, framing CAPTCHAs as tests of spatial intelligence. Yet, as Table~\ref{tab:captcha-compare} shows, gaps remain: Open CaptchaWorld~\citep{luo2025open} uses synthetic data without reasoning labels; Halligan~\citep{teoh2025bot-hard} and OEDIPUS~\citep{deng2024oedipus} provide real data but lack reasoning annotations; and MCA-Bench, though large, is synthetic and detached from real-world challenges. By contrast, our CAPTCHA-X is one of the few large-scale real-world datasets (1,839 puzzles), and uniquely enriched with detailed reasoning annotations and region-based validation. This makes it the first benchmark to evaluate both solving accuracy and reasoning in vision–language models under realistic conditions.

\textbf{Reasoning in Visual CAPTCHA Solving.} Reasoning has become a decisive factor in solving modern CAPTCHAs. Early VLM-based solvers emphasized perceptual accuracy but failed on tasks requiring spatial inference or multi-step logic \citep{shi2019adversarial}. Later work explored adversarial and cognitive-inspired CAPTCHA designs, showing that robustness depends not only on recognition but also on following reasoning chains \citep{bursztein2011text,yan2016survey}. Recent methods employ large language models to guide multi-modal perception, yet their evaluation usually reports only final accuracy without reasoning annotations or ablations \citep{ye2022recognizing}. Platforms like Open CaptchaWorld attempted to capture reasoning complexity with new metrics, but still lacked reasoning annotations.

\textbf{Spatial Reasoning Benchmarks.}
Spatial reasoning is central to visual intelligence, motivating benchmarks such as ARC-AGI \citep{chollet2019measure} with grid-based puzzles testing object permanence and spatial relations, CLEVR \citep{johnson2017clevr} for compositional reasoning, and PTR \citep{hong2021ptr} for part-whole hierarchies. Extending to 3D, 3DSRBench \citep{ma2024_3dsrbench} exposes large human–machine gaps. Distinctly, our CAPTCHA benchmark leverages decades of adversarially tested human and machine challenges, offering spatial reasoning tasks inherently designed to reveal AI weaknesses.

\section{Method}
\subsection{Data Collection and Curation}
\label{sec:3.1}

\begin{figure}[!h]
  \centering
  \includegraphics[width=0.45\textwidth]{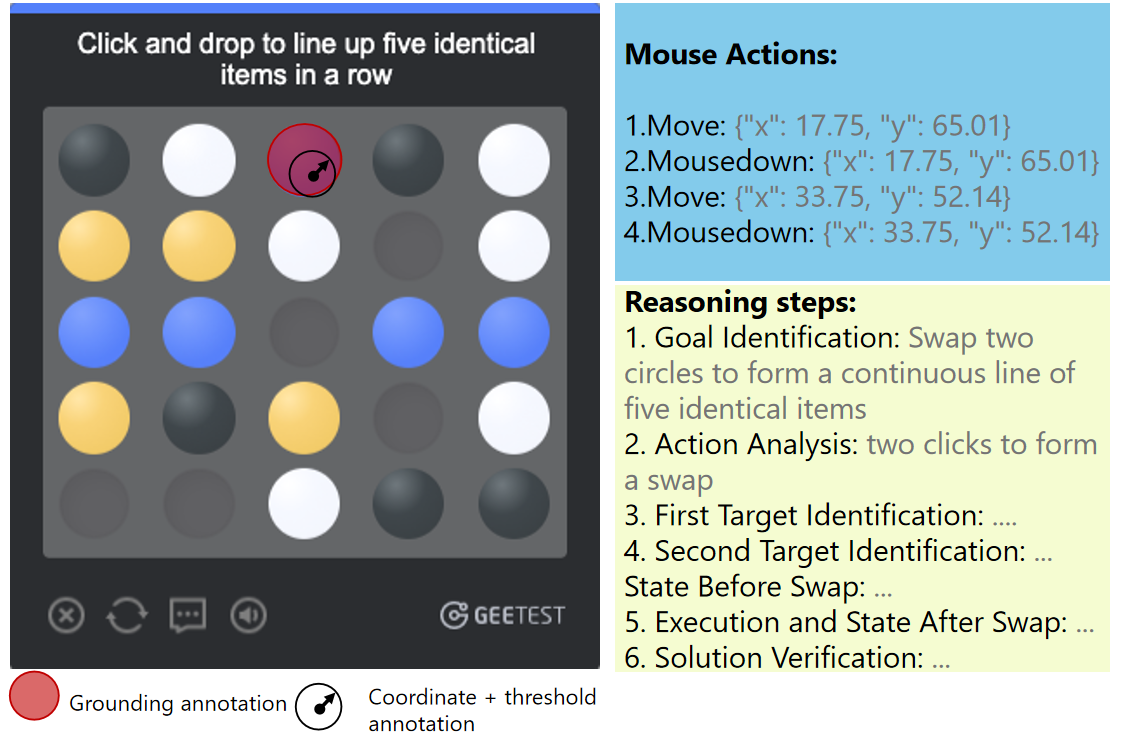}
   \caption{Grounding annotation (red) versus threshold-based annotation (black) in a \textbf{GeeTest Gobang} puzzle, along with recorded mouse actions and reasoning steps. These mouse actions and reasoning steps are generated by using carefully designed prompts.}
   \label{fig:captcha_example}
\end{figure}

To address the limitations of existing benchmarks, we developed \benchmark\ through a systematic data collection pipeline with high-quality,  reasoning steps annotations.

\textbf{Data Collection.} 
We collect CAPTCHA data by programmatically interacting with websites using Selenium~\citep{selenium} and PyAutoGUI~\citep{pyautogui}, while recording comprehensive mouse action sequences and screenshots before and after each puzzle. 

\textbf{Grounding Annotation Generation.}
After solving a CAPTCHA, we record the click coordinates, which may not fall exactly at the object center. We therefore define acceptance regions by manually marking all valid circles or boxes and count a click as correct if it falls within one of them. Unlike prior work that uses a fixed threshold around the click, our approach covers the full target area more reliably, as shown in \autoref{fig:captcha_example}.

\textbf{Reasoning Steps Generation.} 
To obtain both accurate mouse actions and reasoning annotations, we use LLMs (GPT-5) to generate the reasoning steps, and only then derive the mouse click coordinates strictly following these steps. We choose LLM-based generation because manual annotation is highly labor-intensive, and manually written reasoning steps tend to lack diversity.
Concretely, we generate the reasoning steps by employing carefully designed prompts that are (1) goal-directed, clearly describing the task objective and the criteria for identifying the correct regions without revealing any ground-truth click locations, (2) vision-language aware, maximally exploiting the LLM’s ability to jointly process visual content and text, (3) naturally expressed, encouraging concise and conversational reasoning steps, and (4) challenging, designed to maximally elicit the model’s reasoning ability. (5) aligned with established methodology: prior work uses LLM to generate textual annotations for evaluating downstream models (e.g., EgoTextVQA \cite{egotextvqa2025}, EgoTempo \cite{plizzari2025egotempo}, etc.), and we adopt the same general strategy.

\begin{figure}[t]
  \centering
  \includegraphics[width=0.45\textwidth]{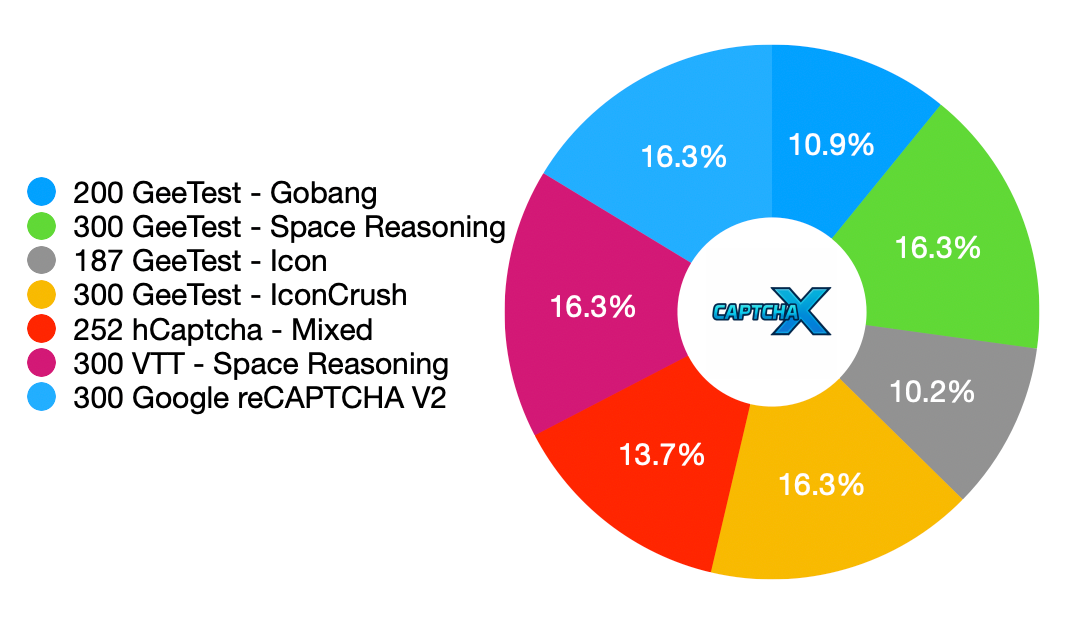}
   \caption{Distribution of our benchmark.}
   \label{fig:dist}
\end{figure}

\textbf{Quality Assurance}
To ensure the reliability and accuracy of \benchmark, every generated reasoning step underwent rigorous human verification by four domain experts. Each expert independently scored the quality of the reasoning steps on a 0–10 scale. If the score difference among the experts exceeded 2 points, or if the average score fell below 5, the sample was jointly re-examined. Expert agreement reached 98\% under this criterion, and the remaining cases were resolved through discussion, yielding 100\% consensus in the final annotations. We conduct extensive experiments and statistical analyses in the Appendix B to validate the effectiveness of our reasoning steps.

\textbf{\benchmark}
Our benchmark comprises 1,839 CAPTCHA puzzles across seven categories, as shown in \autoref{fig:dist}. It covers grid-based puzzles, spatial reasoning tasks, and mixed styles, with each category contributing about 10–16\% of the total for balanced distribution. For every puzzle, we provide reasoning steps and mouse action sequences to evaluate both solving accuracy and reasoning quality. An example from Gobang is shown in \autoref{fig:captcha_example}.

\subsection{CAPTCHA Evaluation Metrics}
To systematically evaluate models' capability in solving CAPTCHAs, we define a comprehensive evaluation metric.
Specifically, our metrics consider both the correctness of actions and the reasoning by comparing with our annotated ground truth.
%
%




We formalize the answer to a CAPTCHA puzzle as an ordered sequence:
\begin{equation}
\mathcal{S} = \{ (a_1, c_1), (a_2, c_2), \dots, (a_m, c_m); \, R \} ,
\end{equation}
where $(a_i, c_i)$ denotes the $i$-th action and its associated coordinate; $R = \langle r_1, r_2, \dots, r_k \rangle$ denotes the reasoning process, expressed as a sequence of steps.


\subsubsection{Action Accuracy} 
Our metric measures if the predicted action and coordinate sequence 
$\{(a_1, c_1), (a_2, c_2), \dots, (a_N, c_N)\}$ exactly matches the ground-truth sequence 
in both order and correctness. 
Let $a_i^\ast$ denote the ground-truth action at step $i$, 
$(\hat{x}_i, \hat{y}_i)$ denote the predicted coordinate $c_i$, 
and $\mathcal{RG}_i$ the corresponding acceptance region. 
We define sequence-level accuracy as: 

\begin{equation}
\begin{split}
AccRate &= \frac{1}{M}\sum_{j=1}^{M} 
\mathbf{1}\Big( a_i^{(j)} = a_i^{\ast (j)} \;\wedge\; \\
&\quad (\hat{x}_i^{(j)},\hat{y}_i^{(j)}) \in \mathcal{RG}_i^{(j)}, \;\forall i \Big)
\end{split}
\end{equation}
where $M$ is the total number of CAPTCHA puzzles. 
Here $\mathbf{1}\{\cdot\}$ returns $1$ only if the entire predicted sequence exactly matches the 
ground truth in both action order and coordinates, and $0$ otherwise.


\subsubsection{Reasoning Accuracy}
To comprehensively evaluate the quality of model-predicted reasoning, we design multiple new metrics for reasoning, each motivated by a distinct aspect of reasoning quality. We argue that high-quality reasoning steps should achieve high solving accuracy or capture maximal complexity with minimal reasoning cost.

\textbf{Reasoning Steps.} To measure the granularity of reasoning, we count the number of reasoning steps in the generated textual reasoning. This metric naturally reflects the level of detail in the reasoning process. A larger number of steps typically implies a more complex reasoning trajectory, but also indicates reduced reasoning efficiency.

\textbf{Reasoning Length.} 
We measure the total number of tokens in the generated reasoning text. In contrast to Reasoning Steps, which capture the structural depth of reasoning, this metric quantifies the overall textual length, offering a finer-grained view of reasoning cost.

\textbf{Reasoning Efficiency.} To assess the trade-off between predictive accuracy and reasoning cost, we define an efficiency metric. Let $Acc_i$ denote the accuracy of model $i$, $\hat{L}_i=L_i/\overline{L}$ the normalized reasoning length, and $\hat{S}_i=S_i/\overline{S}$ the normalized reasoning steps. With equal weights $\alpha=\beta=0.5$, efficiency is computed as
\begin{equation}
    Efficiency_i = \frac{Acc_i}{\alpha \cdot \hat{L}_i + \beta \cdot \hat{S}_i}.
\end{equation}
Values are further using min–max normalized to $(0,1)$. In all, higher reasoning efficiency reflects the model achieving stronger accuracy with fewer steps or tokens, which is more efficient.

\textbf{Trajectory Complexity Index (TCI).} To quantify the structural complexity of reasoning trajectories, we capture linguistic signals such as backtracking words (\emph{but}, \emph{however}, etc.) and symbolic markers (coordinates, grid indices, etc.). For each instance $j$ in group $i$, we aggregate feature counts $F_{i,j}$ and normalize them by group-level averages:
\begin{equation}
    z_{i,j} = \frac{\sum_{F}(F_{i,j} - \overline{F}_i)}{0.5 \cdot (s_i/\overline{s}) + 0.5 \cdot (t_i/\overline{t})}.
\end{equation}
The final TCI is obtained by applying a sigmoid function, which maps the feature values into the normalized range of (0, 1):
\begin{equation}
    TCI_i = \sigma\!\left(\frac{1}{N_i}\sum_{j=1}^{N_i} z_{i,j}\right), \quad \sigma(x) = \frac{1}{1 + e^{-x}}.
\end{equation}
A higher TCI indicates frequent backtracking or symbolic reasoning, demonstrating more complex reasoning behavior.

\textbf{Reasoning Score.} To evaluate how well a model's reasoning aligns with the ground-truth reasoning steps, we use human annotations. Specifically, we recruited four domain experts to independently assess the quality of each generated reasoning sequence across multiple sub-dimensions. For each instance $i$, the final score is obtained by averaging the four expert-provided scores:
\begin{equation}
S_i = \frac{1}{4}\sum_{m=1}^{4} s_{i,m}.
\end{equation}
All experts evaluated the samples in a blinded manner, and disagreements greater than two points triggered a second-round review until consensus was reached. We additionally report inter-rater reliability, with Krippendorff’s alpha of 0.78 across annotators, indicating substantial agreement.

\subsection{Vision-Language Model Agentic Pipeline}

Our agent begins with a \textbf{Category Judger} that routes each puzzle to either a grid-based or a non-grid-based branch. 
This classification is crucial because the two types of puzzles require fundamentally different reasoning strategies. And all the clickable CAPTCHA can be divided into these two categories.
For grid-based puzzles (e.g., Google reCAPTCHA, GeeTest), a dedicated \textbf{Mapping Expert}, implemented as a large language model guided by carefully designed prompts, converts the puzzle board into an $A \times A$ symbolic grid (e.g., $[a,a,a; b,b,c; c,b,b]$). 
This abstraction enables the \textbf{Reasoning Steps Generator} to conduct structured step-by-step inference over the grid, leading to accurate identification of the target cell(s) and high-level planning. 
In contrast, non-grid-based puzzles (e.g., GeeTest Icon, VTT Space Reasoning) rely on spatial semantics rather than grid indexing, and therefore the \textbf{Reasoning Steps Generator} first produces reasoning steps that are refined by a \textbf{Spatial Understanding Expert}, which grounds objects and regions into spatial coordinates.

\begin{figure*}[t]
  \centering
  \includegraphics[width=0.8\textwidth]{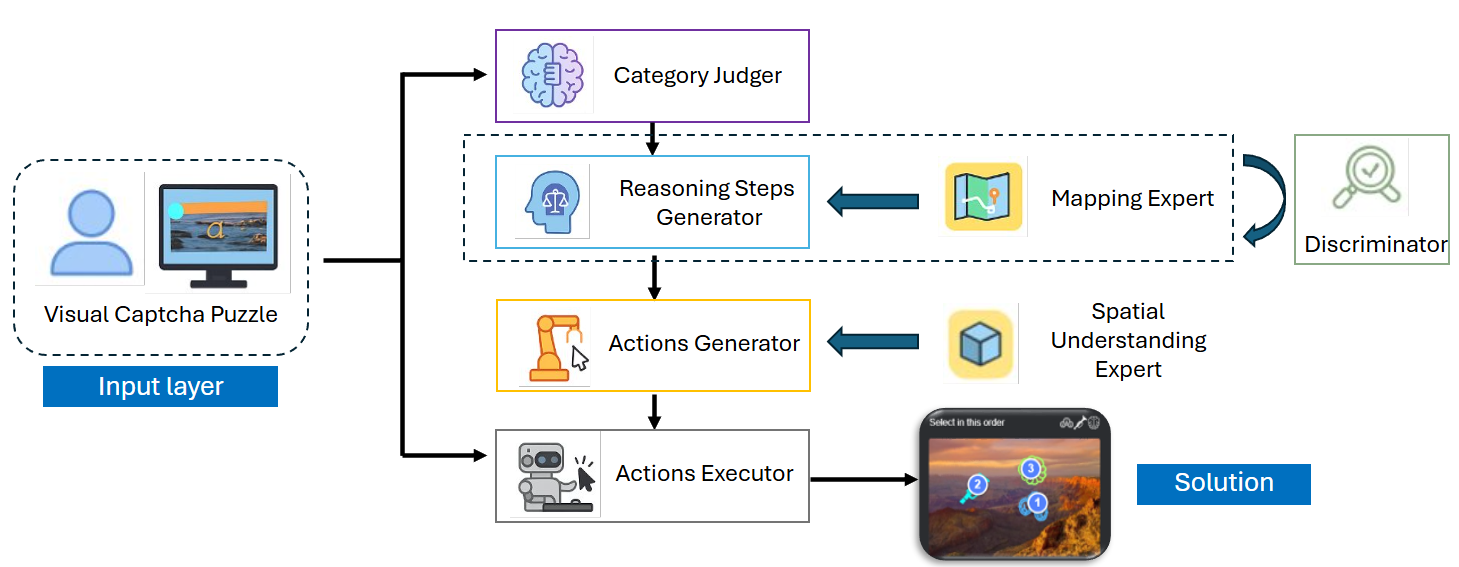}
  \caption{Our Agentic Vision-Language Model  Pipeline.}
  \label{fig:agent_pipeline}
\end{figure*}

To ensure logical consistency across both branches, a \textbf{Discriminator} validates that the generated reasoning is coherent before passing it forward. 
The validated reasoning is then handled by an \textbf{Action Generator}, which translates reasoning outputs into executable click coordinates. Finally, an \textbf{Action Executor} performs the actual clicks on the screen to solve the CAPTCHA. By explicitly distinguishing between grid-based and non-grid-based categories, this unified framework highlights the central role of reasoning in solving diverse visual CAPTCHA.

\section{Experiments}

We conduct experiments to assess the role of reasoning in CAPTCHA solving by comparing model performance with and without reasoning and measuring spatial alignment via $L_2$ distance in a zero-shot setting(reasoning = model-generated reasoning only). All experiments use a fixed API configuration (temperature $=0$, seed $=41$) for reproducibility. We report results in two dimensions: \textbf{Action Evaluation}, which measures end-task accuracy, and \textbf{Reasoning Evaluation}, which analyzes the quality of intermediate reasoning steps. In addition, we include human performance as a reference point. Human results serve as a natural upper bound for CAPTCHA-solving accuracy and provide a meaningful context for interpreting model performance. Specifically, we recruited 20 participants, each of whom independently solved all our benchmark using the same web UI interface provided to the models.
We recorded all trajectories across all participants, which serves as the human baseline.


\subsection{Action Evaluation}

\begin{figure*}[b]
  \centering
  \includegraphics[width=0.9\textwidth]{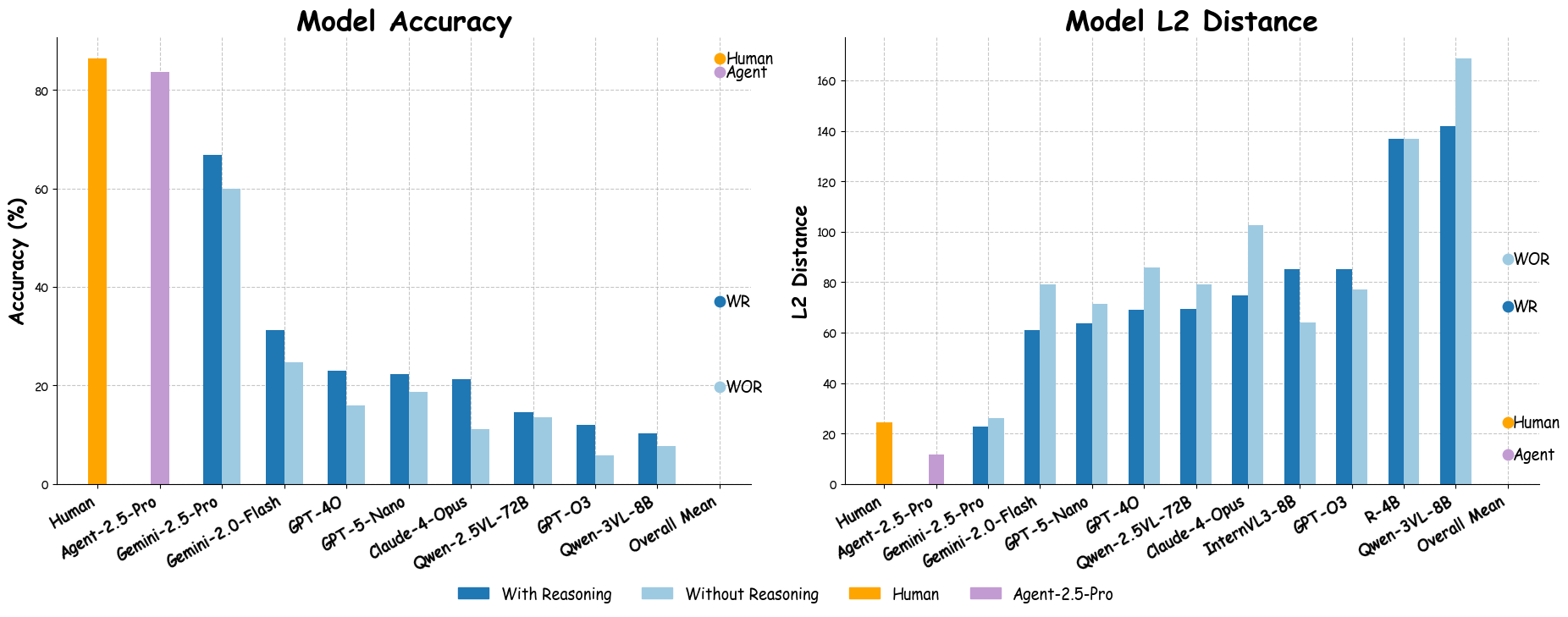}
  \caption{Model Accuracy and L2 Distance with and without reasoning. We averaged over multiple evaluation runs to reduce randomness.
Orange bars indicate Human performance, while purple bars represent our Agent-2.5-Pro's performance. Blue and light-blue bars correspond to model results with reasoning (WR) and without reasoning (WOR), respectively. For accuracy, higher values are better, whereas for L2 distance, lower values are better instead. A clear trend emerges: in nearly all models, WR significantly outperforms WOR. }
  \label{fig:captcha-bar_for_all}
\end{figure*}

\textbf{Evaluation of Prediction Accuracy.}
As shown in ~\autoref{tab:captcha_results}, prompting models to generate reasoning steps almost always improves \textbf{solving accuracy}. Only a small number of cases show marginal decreases ~\autoref{fig:captcha-bar_for_all} illustrates this trend.

\definecolor{lightgreen}{RGB}{220, 255, 220}  

\begin{table*}[ht]
\centering
\scriptsize
\renewcommand{\arraystretch}{1.3} 
\setlength{\tabcolsep}{6pt} 
\caption{Model performance (WR = With Reasoning, WOR = Without Reasoning) across different CAPTCHA types.}
\resizebox{\textwidth}{!}{
\begin{tabular}{c@{\hspace{8pt}}cc@{\hspace{8pt}}cc@{\hspace{8pt}}cc@{\hspace{8pt}}cc@{\hspace{8pt}}cc@{\hspace{8pt}}cc@{\hspace{8pt}}cc}
\hline
\multirow{2}{*}{\textbf{\small Model}} & \multicolumn{2}{c@{\hspace{8pt}}}{\textbf{\small Gobang}} & \multicolumn{2}{c@{\hspace{8pt}}}{\textbf{\small Icon}} & \multicolumn{2}{c@{\hspace{8pt}}}{\textbf{\small Iconcrush}} & \multicolumn{2}{c@{\hspace{8pt}}}{\textbf{\small Recaptcha}} & \multicolumn{2}{c@{\hspace{8pt}}}{\textbf{\small Space Reasoning}} & \multicolumn{2}{c@{\hspace{8pt}}}{\textbf{\small hcaptcha}} & \multicolumn{2}{c}{\textbf{\small VTT}} \\
\cline{2-15}
 & WR & WOR & WR & WOR & WR & WOR & WR & WOR & WR & WOR & WR & WOR & WR & WOR \\
\hline
\noalign{\vskip 2pt}
\rowcolor{lightblue}
\multicolumn{15}{l}{\hspace{6pt}\textit{Closed Source Models}} \\ 
GPT-O3           & 2.00  & 0.00  & 22.00 & 29.79 & 3.67  & 3.67  & 10.67 & 1.82  & 10.00 & 1.50  & 27.67 & 0.00  & 7.00  & 3.67  \\
GPT-4O           & 0.00  & 0.00  & 9.52  & 7.48  & 28.00 & 23.33 & 11.00 & 1.52  & 47.00 & 40.00 & 23.71 & 1.92  & 42.00 & 37.67 \\
GPT-5-Nano       & 0.00  & 0.00  & 0.00  & 0.00  & 28.00 & 23.33 & 8.33  & 2.00  & 31.00 & 32.00 & 58.33 & 40.00 & 30.67 & 32.67 \\
Gemini-2.5-Pro   & 57.00 & 48.00 & 59.30 & 46.30 & 75.00 & 66.67 & 64.00 & 56.52 & 68.00 & 64.67 & 80.95 & 81.35 & 63.00 & 56.00 \\
Gemini-2.0-Flash & 2.00  & 0.00  & 36.33 & 39.67 & 2.33  & 2.00  & 36.33 & 31.67 & 53.00 & 51.00 & 43.21 & 0.79  & 45.67 & 47.67 \\
Claude-4-Opus    & 18.00 & 8.00  & 17.65 & 13.00 & 18.00 & 6.67  & 12.33 & 3.33  & 29.00 & 23.33 & 26.70 & 0.00  & 26.67 & 23.67 \\
\hline
\noalign{\vskip 2pt}
\rowcolor{lightgray}
\multicolumn{15}{l}{\hspace{6pt}\textit{Open Source Models}} \\ 
Qwen-2.5VL-72B   & 0.00  & 0.00  & 0.00  & 0.00  & 6.00  & 5.00  & 14.00 & 0.00  & 24.00 & 27.67 & 38.10 & 36.11 & 19.33 & 26.67 \\
Qwen-3VL-8B         & 0.00  & 0.00  & 0.00  & 0.00  & 0.00  & 0.00  & 51.67 & 43.33 & 0.00  & 0.00  & 19.84 & 9.92  & 0.00  & 0.00  \\
InternVL3-8B     & 0.00  & 0.00  & 0.00  & 0.00  & 0.00  & 0.00  & 0.00  & 0.00  & 0.00  & 0.00  & 0.00  & 0.00  & 0.00  & 0.00  \\
R-4B      & 0.00  & 0.00  & 0.00  & 0.00  & 0.00  & 0.00  & 0.00  & 0.00  & 0.00  & 0.00  & 0.00  & 0.00  & 0.00  & 0.00  \\
\hline
\noalign{\vskip 2pt}
\rowcolor{lightred}
\multicolumn{15}{l}{\hspace{6pt}\textit{Ours}} \\ 
Captcha-X-Agent-O3 (Ours)      & 39.00 & -- & \cellcolor{lightgreen}80.10 & -- & \cellcolor{lightgreen}93.00 & -- & 69.40 & -- & 96.67 & -- & 91.74 & -- & 79.00 & -- \\
Captcha-X-Agent-2.5-Pro (Ours) & \cellcolor{lightgreen}67.44 & -- & 78.60 & -- & 92.33 & -- & \cellcolor{lightgreen}73.00 & -- & \cellcolor{lightgreen}98.67 & -- & \cellcolor{lightgreen}94.44 & -- & \cellcolor{lightgreen}80.67 & -- \\
\hline
\end{tabular}
}
\label{tab:captcha_results}
\end{table*}

Gemini-2.5-Pro achieves the highest accuracy among existing models, with Gemini-2.0-Flash and GPT-5-Nano following at moderate levels. Claude-4-Opus, GPT-4O, GPT-O3, and Qwen-2.5VL-72B also benefit from reasoning, though with lower absolute performance. Building on GPT-O3 and Gemini-2.5-Pro, our agentic pipeline achieves the best accuracy across all CAPTCHA categories.

\textbf{Evaluation of L2 Distance.}
Beyond accuracy, our dataset provides region centers to compute $L_2$ distance between predictions and ground truth. This metric directly measures spatial grounding: smaller distances indicate precise localization. Using both accuracy and $L_2$ distance yields a more reliable measure of solving quality.

Our results show that \textbf{Gemini-2.5-Pro achieves the smallest} $L_2$ \textbf{distance among all existing models}, with Gemini-2.0-Flash also performing well. In contrast, weaker models such as R-4B and Qwen3VL-8B produce very large spatial errors, often exceeding 100 pixels. Importantly, our agent obtains the lowest $L_2$ distance on every CAPTCHA type (full numbers are reported in the appendix), even surpassing human performance. These findings demonstrate that $L_2$ distance provides complementary evidence of spatial grounding beyond accuracy alone, as shown in ~\autoref{fig:captcha-bar_for_all}.

\textbf{Statistical Validation.}
To validate whether the $L_2$ distance is a meaningful indicator of spatial reasoning ability, we first examine its relationship with solving accuracy. Our motivation is straightforward: if a model solves more CAPTCHAs correctly, it should also produce more accurate spatial localization. Therefore, we compute each model’s average accuracy and average $L_2$ distance across all CAPTCHA types and perform a regression analysis. The results confirm a clear pattern: models with higher solving accuracy consistently achieve smaller $L_2$ distances. The regression yields a strong negative correlation ($R^2 = 0.97$, $p < 0.001$), and no models deviate from this trend, confirming that $L_2$ is a valid measure of localization performance.

To further assess the impact of reasoning, we apply McNemar’s test~\citep{mcnemar1947note} for paired accuracy outcomes and observe a significant improvement ($p < 0.001$). For spatial error, the Wilcoxon signed-rank test also indicates a significant reduction ($p < 0.001$). Quantitatively, when reasoning increases solving accuracy by 38.75\%, the average $L_2$ distance decreases by 14.6\%. These results provide strong statistical evidence that reasoning not only improves task success but also significantly enhances spatial localization.

 \subsection{Reasoning Evaluation}

\begin{figure}[!h]
  \centering
  \includegraphics[width=0.45\textwidth]{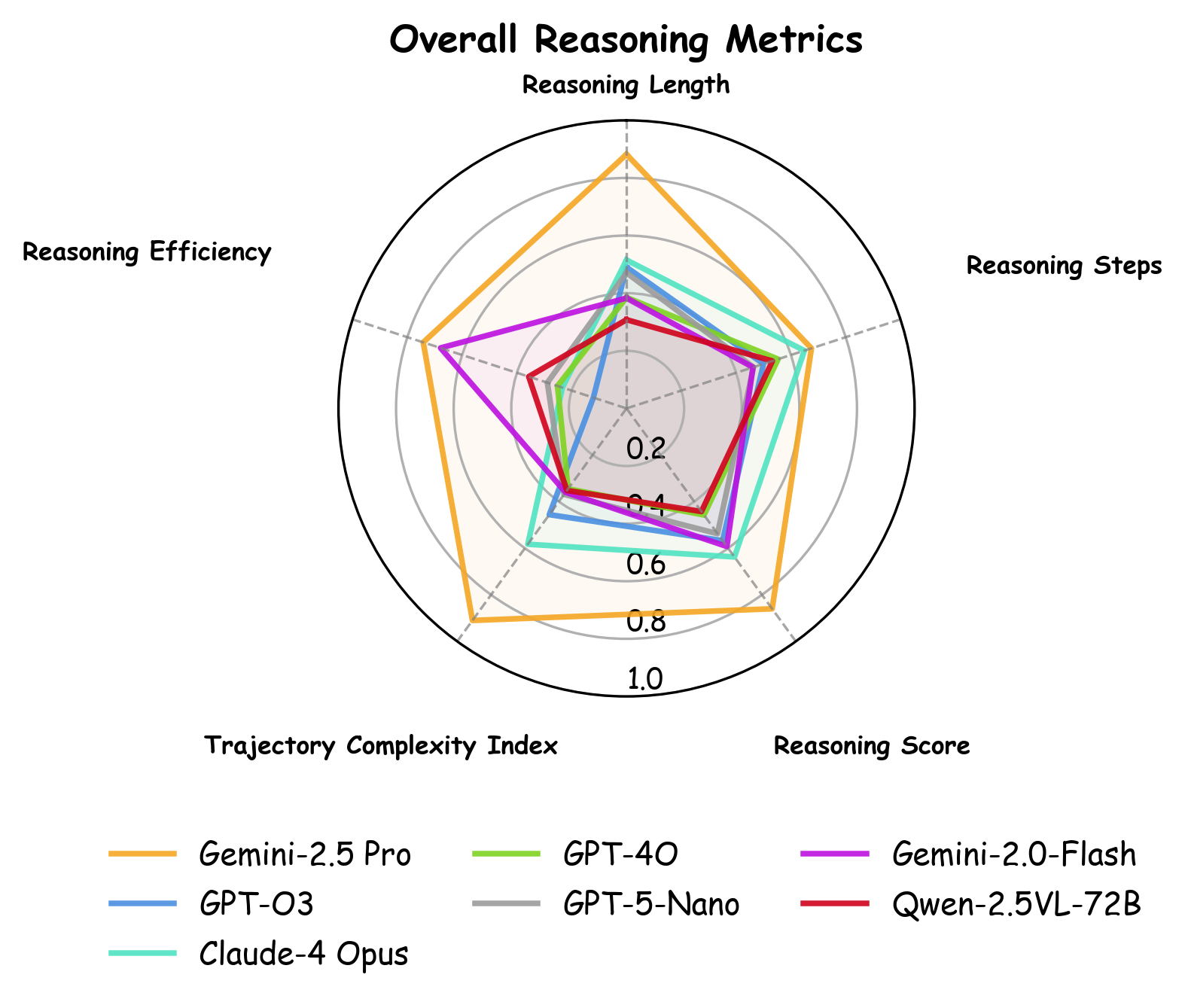}
   \caption{Reasoning Evaluation with Multi-Dimensions: 
  The radar chart shows overall reasoning metrics averaged across CAPTCHA categories.}
   \label{fig:bar_only_rader}
\end{figure}

To systematically assess reasoning quality, we evaluate multiple reasoning metrics here.
\autoref{fig:bar_only_rader}
presents the radar chart: it aggregates overall reasoning metrics averaged across all CAPTCHA categories. Due to space limits, we show representative models in the main figure.

\textbf{Overall Reasoning Metrics.} The radar chart summarizes average reasoning behaviors across models. Gemini-2.5-Pro dominates all dimensions, combining the longest and most information-dense reasoning with the highest efficiency. Claude-4-Opus ranks second in reasoning quality and complexity, but suffers from very low efficiency, ranking near the bottom among all models. In contrast, Gemini-2.0-Flash achieves high efficiency with much shorter reasoning, demonstrating a different trade-off strategy. Weaker models show varied patterns. GPT-O3 has the lowest efficiency despite moderate reasoning length. Qwen-2.5VL-72B produces the shortest reasoning with the lowest score but maintains mid-tier efficiency. GPT-4O and GPT-5-Nano fall in between with moderate performance across all metrics.

\begin{figure}[!h]
  \centering
  \includegraphics[width=0.4\textwidth]{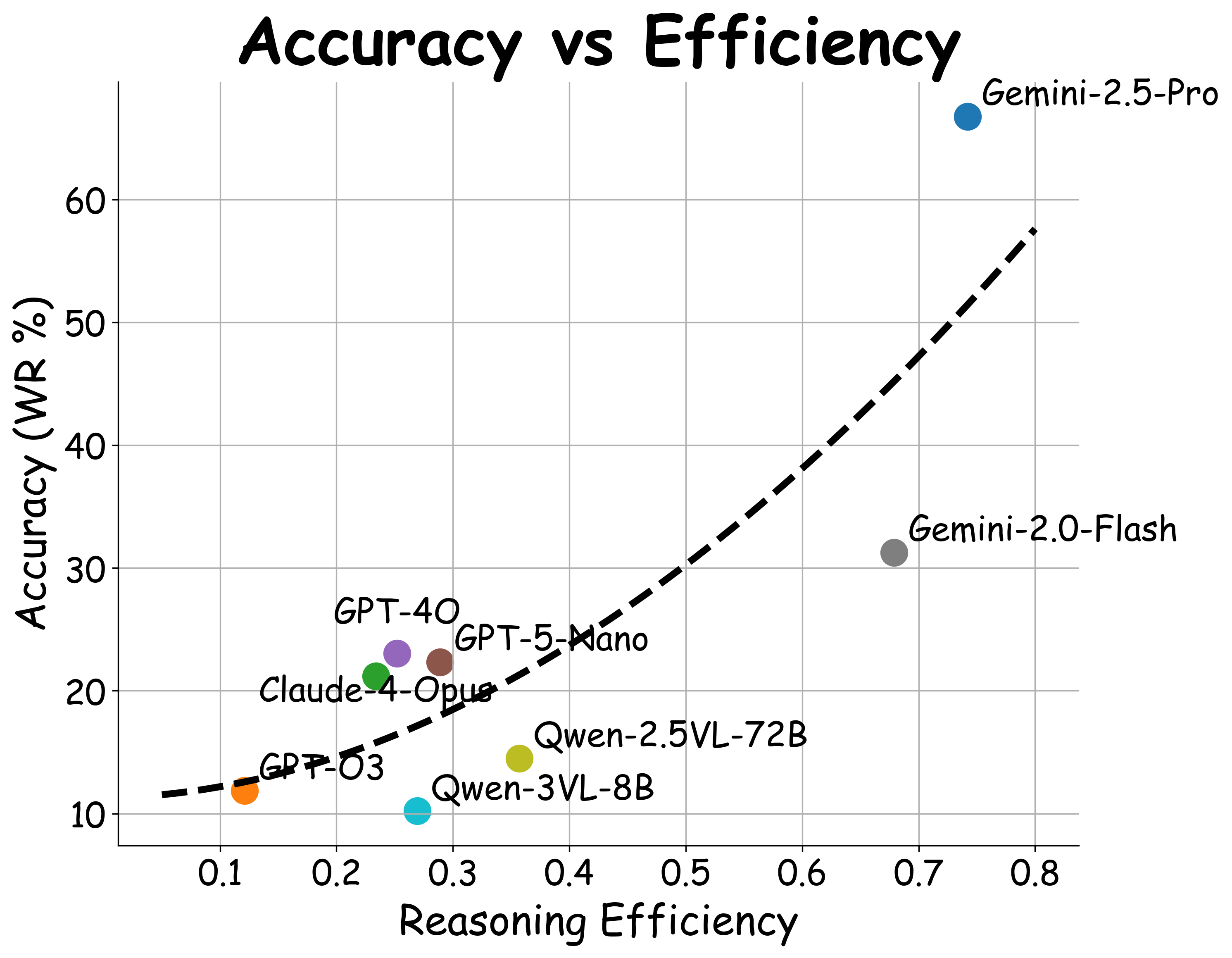}
   \caption{Reasoning Scaling Law.}
   \label{fig:scaling}
\end{figure}

\paragraph{Reasoning Scaling Laws.}
Within our dataset, we observe three empirical \textbf{reasoning scaling laws} patterns across evaluated models.

\textbf{(1) Linear Scaling Law.}
We find a near-perfect linear relationship between reasoning score, reasoning length, and trajectory complexity index (TCI):
\begin{equation}
\begin{aligned}
\text{Reasoning Length} &= a_1 \cdot \text{Score} + b_1, \\
\text{TCI} &= a_2 \cdot \text{Score} + b_2,
\end{aligned}
\end{equation}

where $a_1, b_1, a_2, b_2$ are regression coefficients. 
Empirically, $a_1 = 78.95$, $b_1 = -62.11$, $a_2 = 0.349$, and $b_2 = -0.333$, all significant at $p < 0.01$. 
Since reasoning score strongly predicts accuracy ($r = 0.88$), this law establishes a principled connection between reasoning cost and problem-solving ability, enabling accuracy to be directly forecasted from reasoning complexity.

\begin{table*}[h]
\centering
\scriptsize
\renewcommand{\arraystretch}{1.3}
\setlength{\tabcolsep}{6pt} 
\caption{Comparison of CAPTCHA Solving Accuracy for Different CAPTCHA Solvers}
\label{tab:agent}
\resizebox{\linewidth}{!}{
\begin{tabular}{c@{\hspace{8pt}}c@{\hspace{8pt}}c@{\hspace{8pt}}c@{\hspace{8pt}}c@{\hspace{8pt}}c@{\hspace{8pt}}c@{\hspace{8pt}}c@{\hspace{8pt}}c@{\hspace{8pt}}c@{\hspace{8pt}}c}
\hline
\textbf{\small Model} & \cellcolor{lightpurple}\textbf{\small Icon} & \cellcolor{lightpurple}\textbf{\small Space Reasoning} & \cellcolor{lightpurple}\textbf{\small VTT} & \cellcolor{lightpurple}\textbf{\small Iconcrush} & \cellcolor{lightpurple}\textbf{\small hCaptcha} & \cellcolor{lightpurple}\textbf{\small Gobang} & \cellcolor{lightpurple}\textbf{\small Recaptcha} & \cellcolor{orange!20}\textbf{\small Geometry} & \cellcolor{orange!20}\textbf{\small Click Order} & \cellcolor{orange!20}\textbf{\small Animal} \\
\hline
\noalign{\vskip 2pt}
\rowcolor{lightblue}
\multicolumn{11}{l}{\hspace{6pt}\textit{Baseline Models}} \\ 
Baseline        & 46.3 & 64.67 & 50.00 & 66.7 & 0 & 48 & 56.52 & 60 & 55 & 43.33 \\
OEDIPUS-DSL~\citep{deng2024oedipus}     & --   & 65.4  & --    & 67.4 & --   & 80.2 & -- & -- & -- & -- \\
Halligan ~\citep{teoh2025bot-hard}       & 46   & --    & 23    & \cellcolor{lightgreen}98 & 82 & \cellcolor{lightgreen}92 & 68 & -- & -- & -- \\
VTTsolver~\citep{gao2021vttsolver}       & --   & 90.8    & 50    & --   & --   & --   & -- & -- & -- & -- \\
PhishDecloaker~\citep{teoh2024phishdecloaker}  & --   & --    & --    & --   & 74   & --   & 72 & -- & -- & -- \\
\hline
\noalign{\vskip 2pt}
\rowcolor{lightred}
\multicolumn{11}{l}{\hspace{6pt}\textit{Ours}} \\ 
Captcha-X-Agent (Ours) & \cellcolor{lightgreen}80.1 & \cellcolor{lightgreen}98.67 & \cellcolor{lightgreen}80.67 & 93 & \cellcolor{lightgreen}94.44 & 67.44 & \cellcolor{lightgreen}73 & \cellcolor{lightgreen}100 & \cellcolor{lightgreen}85 & \cellcolor{lightgreen}90 \\
\hline
\end{tabular}}
\end{table*}

\textbf{(2) Power Law.}
We futher observe that accuracy exhibits a clear power-law relationship with the proposed reasoning-efficiency metric, which can be well fitted by
\begin{equation}
\text{Accuracy} = A + k \cdot \text{Eff}^{\alpha},
\end{equation}
where $A$ denotes the offset, $k$ the scaling coefficient, and $\alpha$ the exponent.
Across all evaluated models, the fitted parameters are $A = 11.30$, $k = 70.76$, and $\alpha = 1.90$, yielding $R^2 = 0.828$ ($p < 0.01$).
This power-law fit suggests a superlinear trend in which models with higher reasoning efficiency tend to achieve disproportionately higher accuracy, as illustrated in \autoref{fig:scaling}.

\textbf{(3) Difficulty–Gain Scaling Law.}
Finally, we examine how the benefit of reasoning varies with task difficulty, and we have observed that reasoning becomes more effective as tasks grow harder. 
To quantify the difficulty level of each CAPTCHA type, we define a  difficulty level as
\begin{equation}
\text{Difficulty} = 1 - \overline{\text{Acc}},
\end{equation}

where $\overline{\text{Acc}}$ is the average accuracy achieved by our different vision-language models on that CAPTCHA type. 
This metric reflects the average “solving difficulty” observed across models. When models achieve low accuracy, it indicates that this type of task is hard to solve.

To fairly measure how reasoning contributes under varying levels of difficulty, we adopt the \emph{relative gain} rather than the absolute improvement. Relative gain directly captures how much the reasoning-enhanced performance improves over the non-reasoning baseline.
Formally, we compute:
\begin{equation}
\text{Relative Gain} = 
\frac{\text{WR} - \text{WOR}}{\text{WOR}} \times 100\%.
\end{equation}

We observed that the relative gain exhibits a clear positive correlation with task difficulty. 
It is clear that CAPTCHA types with high difficulty yield substantially larger improvements from reasoning, while easier CAPTCHA types benefit only marginally. 
A linear regression gives $R^{2} = 0.85$ ($p < 0.01$), and Spearman’s rank correlation yields $\rho = 0.93$ ($p = 0.0025$), confirming a strong and statistically significant monotonic trend.

These findings empirically validate our proposed observations again: reasoning becomes increasingly effective as the problem difficulty rises.  This naturally leads to a central question: why does reasoning become more effective as tasks grow harder?

\textbf{(4) Why is reasoning more effective on harder tasks? }  
To further understand why reasoning becomes more beneficial on harder tasks,  
we analyze the relationship between task difficulty and the average reasoning length used by the model.  
Empirically, we observe a strong linear trend between the two quantities:
\begin{equation}
L_{\text{reason}} = \alpha \cdot D_{\text{task}} + \beta, \quad R^2 = 0.92,
\end{equation}
where $L_{\text{reason}}$ denotes the average reasoning length and $D_{\text{task}}$ represents the different Captcha-Solving task difficulty.  we also obtain that $\alpha = 211.71$ and $\beta = -39.95$.
This result shows that as task difficulty increases, models tend to generate longer reasoning sequences. These observations suggest that reasoning implicitly adapts computational resources according to task complexity.  
Harder tasks are associated with proportionally greater reasoning effort, which leads to higher performance gains.  

\subsection{Failure Case Analysis}
Across our experiments, we also observe three major classes of failure. 
\textbf{(1) Logical errors}: the model produces incoherent or self-contradictory reasoning steps. 
\textbf{(2) Mapping errors}: the model misinterprets the underlying grid structure (e.g., treating a 5$\times$5 Gobang CAPTCHA as 3$\times$3 or 8$\times$8). 
\textbf{(3) Grounding errors}: even when the reasoning logic is correct, the model fails to output an accurate final coordinate. 
For a more detailed analysis of these failure modes, please refer to Appendix D.

\subsection{Agentic Evaluation}

To address the failure modes identified above, we introduce our reasoning-centric agentic pipeline. Inspired by the SayCan ~\citep{saycan2022arxiv} in robotics, we propose the \textbf{Discriminator} mitigates logical errors by filtering incoherent reasoning, the \textbf{Mapping Expert} corrects grid-structure misunderstandings, and the \textbf{Spatial Understanding Expert} refine spatial localization to eliminate grounding errors. This design not only substantially improves performance over the direct-prediction baseline but also effectively resolves the major error types, surpassing a wide range of existing CAPTCHA solvers. The ablation study of our agent pipeline is in Appendix, which evaluates individual contributions of each module.

We evaluate both a direct-prediction baseline and our proposed reasoning-centric agentic pipeline for CAPTCHA solving. The baseline uses Gemini-2.5-Pro without reasoning, where the model directly outputs click coordinates from the CAPTCHA image. For broader comparison, we also include the only two general-purpose agent systems available in Captcha: Halligan and OEDIPUS. 
In contrast to these methods, our agent achieves state-of-the-art performance on five of the seven tasks (\autoref{tab:agent}), while also remaining competitive on Iconcrush (93) and Gobang (67.44).

To assess generalization, we further test our agent on three external CAPTCHA types from the Open CaptchaWorld~\citep{luo2025open} benchmark(the orange section in the table). These tasks are not part of our dataset and therefore represent a clear out of distribution setting. Our agent achieves 100 percent on Geometry, 85 percent on Click Order, and 90 percent on Animal, while the benchmark itself reports a highest average accuracy of only 40 percent. These results show that our agent enables strong transfer across unseen CAPTCHA formats.

\section{Discussion}
CAPTCHA-X is developed to study multimodal visual–spatial reasoning on CAPTCHA-like tasks, not to bypass real authentication systems. Only processed static images and anonymized action annotations will be released, without any interaction scripts or website-specific metadata. The benchmark contains no user or system identifiers and is intended solely for research on the model's reasoning capability, encouraging responsible and non-operational use.

\section{Conclusion}
Our work shows that reasoning is a decisive capability for solving modern visual CAPTCHA. With \benchmark, we pair real-world CAPTCHA challenges with reasoning steps, introduce reasoning-oriented metrics, and propose an agentic pipeline that isolates the role of reasoning. These findings highlight reasoning as central to advancing multimodal AI in the future .
{
    \small
    \bibliographystyle{ieeenat_fullname}
    \bibliography{main}
}
\end{document}